\documentclass[journal]{IEEEtranTIE}
\renewcommand{\baselinestretch}{0.9}
\pdfoutput=1
\usepackage{graphicx}
\usepackage{cite}
\usepackage{picinpar}
\usepackage{url}
\usepackage{flushend}
\usepackage[latin1]{inputenc}
\usepackage{colortbl}
\usepackage{soul}
\usepackage{multirow}
\usepackage{pifont}
\usepackage{color}
\usepackage{alltt}
\usepackage[hidelinks]{hyperref}
\usepackage{enumerate}
\usepackage{siunitx}
\usepackage{breakurl}
\usepackage{epstopdf}
\usepackage{pbox}
\usepackage[cmex10]{amsmath}
\usepackage{amsfonts}
\usepackage{bm}
\usepackage{amssymb}  
\usepackage{booktabs}
\usepackage{subfig}
\usepackage{tabularx}
\usepackage{multicol}

\newtheorem{property}{Property}
\newtheorem{theorem}{Theorem}
\newtheorem{assumption}{Assumption}

\begin{document}

\setlength{\lineskiplimit}{0pt}
\setlength{\lineskip}{0pt}
\setlength{\abovedisplayskip}{3pt}   
\setlength{\belowdisplayskip}{3pt}
\setlength{\abovedisplayshortskip}{3pt}
\setlength{\belowdisplayshortskip}{3pt} 

\title{Adaptive RISE Control for Dual-Arm Unmanned Aerial Manipulator Systems with Deep Neural Networks}

\author{
	\vskip 1em
    Yang Wang,
	Hai Yu, \emph{Student Member,~IEEE},
    Shizhen Wu, \emph{Student Member,~IEEE},
    Zhichao Yang,
	\\ Jianda Han, \emph{Member,~IEEE},
    Yongchun Fang, \emph{Senior Member,~IEEE},
    and Xiao Liang, \emph{Member,~IEEE}
    \thanks{
      This work was supported in part by the National Natural Science Foundation of China under Grant 62273187, Grant 62233011, and Grant 623B2054, in part by the Natural Science Foundation of Tianjin under Grant 23JCQNJC01930, and in part by the Haihe Laboratory of ITAI under Grant 22HHXCJC00003. \emph{(Corresponding author: Xiao Liang)}

    The authors are with the Institute of Robotics and Automatic Information System, College of Artificial Intelligence, and Tianjin Key Laboratory of Intelligent Robotics, Nankai University, Tianjin 300350, China, and also with the Engineering Research Center of Trusted Behavior Intelligence, Ministry of Education, Nankai University, Tianjin 300350, China (e-mail: \url{wangy1893@mail.nankai.edu.cn}; \url{yuhai@mail.nankai.edu.cn}; \url{szwu@mail.nankai.edu.cn}; \url{yangzc@mail.nankai.edu.cn}; \url{hanjianda@nankai.edu.cn}; \url{fangyc@nankai.edu.cn}; \url{liangx@nankai.edu.cn}).
    }
}

\maketitle

\pagestyle{empty}       
\thispagestyle{empty}

\begin{abstract}
The unmanned aerial manipulator system, consisting of a multirotor UAV (unmanned aerial vehicle) and a manipulator, has attracted considerable interest from researchers. Nevertheless, the operation of a dual-arm manipulator poses a dynamic challenge, as the CoM (center of mass) of the system changes with manipulator movement, potentially impacting the multirotor UAV. Additionally, unmodeled effects, parameter uncertainties, and external disturbances can significantly degrade control performance, leading to unforeseen dangers. To tackle these issues, this paper proposes a nonlinear adaptive RISE (robust integral of the sign of the error) controller based on DNN (deep neural network). The first step involves establishing the kinematic and dynamic model of the dual-arm aerial manipulator. Subsequently, the adaptive RISE controller is proposed with a DNN feedforward term to effectively address both internal and external challenges. By employing Lyapunov techniques, the asymptotic convergence of the tracking error signals are guaranteed rigorously. Notably, this paper marks a pioneering effort by presenting the first DNN-based adaptive RISE controller design accompanied by a comprehensive stability analysis. To validate the practicality and robustness of the proposed control approach, several groups of actual hardware experiments are conducted. The results confirm the efficacy of the developed methodology in handling real-world scenarios, thereby offering valuable insights into the performance of the dual-arm aerial manipulator system.
\end{abstract}


\markboth{IEEE TRANSACTIONS ON INDUSTRIAL ELECTRONICS}%
{}

\definecolor{limegreen}{rgb}{0.2, 0.8, 0.2}
\definecolor{forestgreen}{rgb}{0.13, 0.55, 0.13}
\definecolor{greenhtml}{rgb}{0.0, 0.5, 0.0}

\section{Introduction}

\IEEEPARstart{I}{n} recent times, the realm of robotics has witnessed a surge in interest in the coordinated operation of dual-arm robots \cite{dualarm1, dualarm2, dualarm3}, highlighting their superior performance compared to single-arm counterparts. The versatile manipulation abilities and extensive workspaces demonstrated by dual-arm robots have ignited a fresh wave of exploration into their integration with multirotor UAVs, which harnesses the dual-arm's extensive workspace while leveraging the distinctive features of multirotor UAVs, such as vertical takeoff, landing capabilities, and enhanced flexibility \cite{chara1, chara2}. In the meantime, the integration of aerial manipulator systems expands the scope of UAV applications beyond passive scenarios \cite{passive2, passive3}, such as search and rescue, monitoring, and surveillance, venturing into active domains \cite{passive4, active4}, including transportation, assembly, and object manipulation. 

Over the past few years, there has been a growing body of research dedicated to the exploration and development of the aerial manipulator systems. Zhong \emph{et al}. \cite{singlezh} focus on the autonomous control of an unmanned aerial manipulator using computer vision technology to grasp target objects. Chen \emph{et al}. \cite{singlecyj} propose a finite-time control strategy employing an adaptive sliding-mode disturbance observer (ASMDO) for an unmanned aerial manipulator to handle uncertainties and external disturbances effectively. Kim \emph{et al}. \cite{singlekim} use an aerila manipulator to open and close an unknown drawer by analyzing the interaction. The developments of unmanned aerial manipulator systems in control, teleoperation, perception, and planning are detailed in \cite{singleollero}.

Limited by the workspaces and capabilities of the single arm, the dual-arm aerial manipulator systems have been developed rapidly. Noteworthy contributions in this field include the continuous research efforts undertaken by Suarez \emph{et al}. \cite{suarez5}. Their works encompass the design of a lightweight and compliant dual-arm manipulator structure, accompanied by rigorous testing and verification processes applied to such scenarios as power lines and pipelines. For the specific application of valve rotation, Orsag \emph{et al}. \cite{orsag1} introduce a dual-arm aerial manipulator system endowed with multiple degrees of freedom. A novel image-based visual-impedance control law is presented by Lippiello \emph{et al}. \cite{app1}, which facilitates the physical interaction of a dual-arm unmanned aerial manipulator equipped with both a camera and a force/torque sensor. Kong \emph{et al}. \cite{app2} contribute to the field by introducing the mechanical and control design of an innovative teleoperated dual-arm aerial platform. The arms of this platform are characterized by a unique joint structure involving tendons and are supported by elastic components. Additionally, Yang \emph{et al}. \cite{app3} focus on the development of a novel dual-arm aerial manipulator system, which is characterized by low weight, low inertia, and humanoid arm structure, emphasizing flexibility in operation.

The primary challenge faced by the dual-arm unmanned aerial manipulator system stems from the inherent impact of arm movements on the multirotor. These impacts cannot be directly compensated for due to the presence of unmeasured signals related to joint velocity and acceleration. Additionally, real-world factors such as unmodeled effects and external disturbances pose unexpected risks to the system. To address these issues, RISE feedback controller could be considered, which is presented in \cite{rise1, rise2, rise4}. However, the RISE method, being a high-gain feedback tool, motivates the incorporation of a feedforward control element alongside the feedback controller. This combination is aimed at achieving potential benefits, including enhanced transient and steady-state performance, as well as reduced control efforts \cite{rise&nn1}. In general, the feedforward component is typically chosen as neural networks (NNs), recognized for their capacity as universal function approximators capable of modeling continuous functions across a compact domain. Shin \emph{et al}. \cite{rise&nn2} develop the position tracking control system of a multirotor UAV by incorporating RISE feedback and a NN feedforward term. While conventional NN-based adaptive controllers guarantee stability with uniformly ultimately bounded, the novel NN-based adaptive control system proposed in this study ensures semi-global asymptotic tracking for the UAV by leveraging the RISE feedback control strategy.

Although NNs with a single hidden layer can approximate general nonlinear functions, the utilization of DNNs is acknowledged for providing superior performance. The outcomes presented in \cite{DNN1, DNN2} leverage offline DNN training methods to approximate explicit model predictive control laws. Nonetheless, the application of such offline approaches comes with inherent limitations, given the substantial data requirements for training and the implementation of derived feedforward terms as open-loop approximators. In contrast to the aforementioned offline training methods, an alternative approach involves the derivation of NN weight update laws based on Lyapunov-based stability analysis, as outlined in \cite{ANN}, enabling the real-time adjustment of NN weights. However, the NN weights are embedded within activation functions, rendering the derivation of adaptation laws from stability analysis particularly challenging when confronted with neural networks comprising more than one hidden layer.

Given the the aforementioned challenges, this research establishes a comprehensive kinematic and dynamic model for a dual-arm aerial manipulator system. To address the inherent issues arising from arm movements and external disturbances, we introduce an innovative approach, i.e., the DNN-based adaptive RISE controller. The primary contributions of this study are outlined as follows:
\begin{enumerate}
\item In practical scenarios, dual-arm unmanned aerial manipulator systems encounter inevitable challenges, including the impact of manipulator actions on the multirotor, uncertainties in parameters, and external disturbances. These factors can significantly degrade control performance. Therefore, this paper combines deep neural networks with the adaptive RISE controller to systematically tackle these issues, ensuring the effective mitigation of UAV tracking errors. In addition to theoretical analysis, the proposed control scheme's effectiveness and robustness are validated through hardware experiments. These experiments serve as practical demonstrations of the system's capability to handle real-world challenges and uncertainties.
\item Utilizing DNNs can yield superior performance compared to single-layer NNs. Furthermore, to tackle the complexities arising from nested nonlinear parameterizations of the inner-layer DNN weights, a recursive representation of the inner-layer DNN structure is introduced to facilitate the analysis process. By using Lyapunov techiniques, it is proven that the tracking error can converge asymptotically. Notably, it presents pioneering outcomes with real-time weight adaptation laws based on RISE for each layer of the DNN.
\end{enumerate}

The paper follows a structured organization outlined as follows. Section \ref{sec:mod} establishes the model of the system, providing a foundational basis for the subsequent content. Section \ref{sec:controller} details the procedure for controller development. Subsequently, Section \ref{sec:stability} develops the stability analysis. In Section \ref{sec:exp}, a self-built dual-arm aerial manipulator system is employed to experimentally verify the proposed methodology. The paper concludes with Section \ref{sec:con}, presenting both conclusions and future works.

\section{Modeling} \label{sec:mod}
\subsection{Kinematic model}
\begin{figure}[t]
\centering
\includegraphics[width=3.0in]{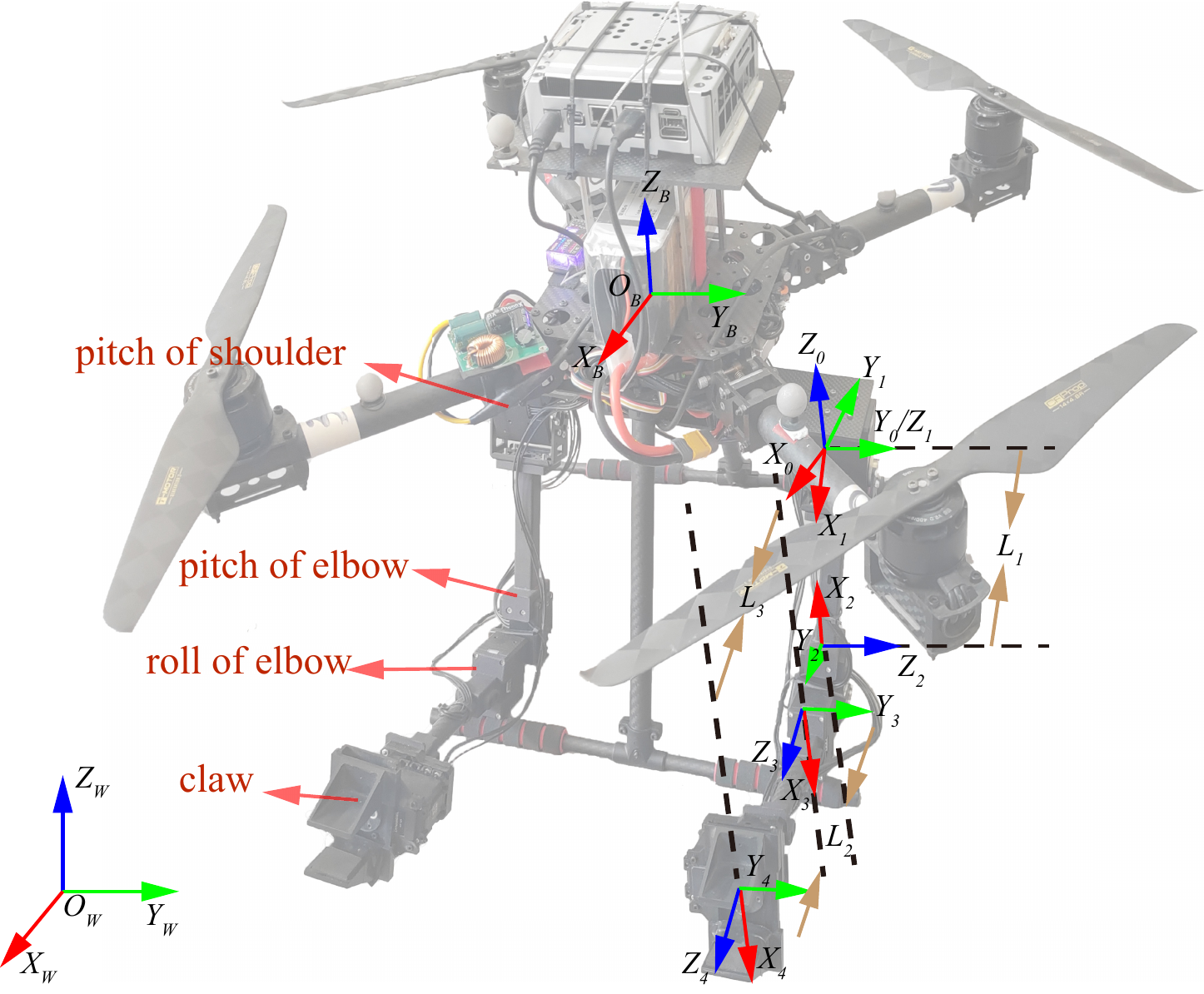}
\caption{Illustration of dual-arm aerial manipulator system.}
\label{fig:dualarm}
\vspace{-0.6cm}
\end{figure}

The diagram of the dual-arm unmanned aerial manipulator system is shown in Fig. \ref{fig:dualarm}. The coordinate frames of the system include the right-hand inertia coordinate frame $\mathcal{F}_W$, the multirotor body-fixed coordinate frame $\mathcal{F}_B$, and the joint frames $\mathcal{F}_i^j(X_i^j$-$Y_i^j$-$Z_i^j)$ of the dual-arm manipulator with $i=0, 1, 2, 3, 4$, where $j=1, 2$ denotes the right and the left arm, respectively. The link length is represented by $L_i (i = 1, 2, 3)$. Each arm possesses three degrees of freedom: pitch of the shoulder, pitch of the elbow, and roll of the elbow, in addition to a claw. Taking the right arm as an example, the forward kinematics of the dual-arm manipulator are derived utilizing DH parameters.

Let $\bm{p}_e^1 \in \mathbb{R}^3, {\Phi}_e^1 \in \mathbb{R}^{3\times3}$ represent the position and orientation of the manipulator's right end-effector with respect to $\mathcal{F}_W$, which are related to the pose of the multirotor as follows:
\begin{align}
\label{DNN:mk}
\left\{
\begin{aligned}
\bm{p}_e^1 &= \bm{p} + {R}\, ^B{\bm{p}}_e^1, \\
{\Phi}_e^1 &= {R}\, ^B{{\Phi}}_e^1,
\end{aligned}
\right.
\end{align}
where $\bm p =\left[p_{x}, p_{y}, p_{z} \right]^\top \in \mathbb R^3$ is the position of the multirotor with respect to $\mathcal F_W$, $R \in SO(3)$ is the rotation matrix from the body-fixed frame $\mathcal{F}_B$ to the inertial frame $\mathcal{F}_W$, and $^B\bm{p}_e^1 \in \mathbb R^3, ^B\!{\Phi}_e^1 \in \mathbb R^{3 \times 3}$ are the position and orientation of the manipulator's end-effector relative to the multirotor with respect to $\mathcal{F}_B$. The velocity relationship between the multirotor and the manipulator's right end-effector can be expressed as follows:
\begin{align}
\left\{
\begin{aligned}
\bm{v}_e^1 &= \bm{v} + {R} ( \bm{\omega} \times ^B\!{\bm p}_e^1 + ^B\!{\bm{v}}_e^1), \\
\bm{\omega}_e^1 &= {R} (\bm \omega + ^B\!{\bm{\omega}}_e^1),
\end{aligned}
\right.
\end{align}
where $\bm{v}_e^1, \bm{\omega}_e^1 \in \mathbb{R}^3$ represent the velocity and angular velocity of the manipulator's right end-effector with respect to $\mathcal{F}_W$, respectively, $\bm v \in \mathbb{R}^3$ represents the velocity of the multirotor with respect to $\mathcal{F}_W$, and $\bm \omega \in \mathbb{R}^3$ represents the angular velocity of the multirotor with respect to $\mathcal{F}_B$. ${^B \bm{v}}_e^1, {^B\bm{\omega}}_e^1 \in \mathbb R^3$ are the velocity and angular velocity of the manipulator's right end-effector relative to the multirotor with respect to $\mathcal{F}_B$, which are related to the angular velocity of the manipulator joint as follows:
\begin{align}
\begin{bmatrix}
{^B\bm{v}}_e^1 \\
{^B\bm{\omega}}_e^1
\end{bmatrix}
= J(\bm \eta^1) \dot{\bm \eta}^1,
\end{align}
where $\bm \eta^1 = [\eta_1^1, \eta_2^1, \eta_3^1]^\top \in \mathbb R^3$ denotes the joint position vector and $J(\bm \eta^1) \in \mathbb R^{6 \times 3}$ denotes the manipulator Jacobian matrix.
\subsection{Dynamic model}
Combining with the Linear Momentum Theorem, the dynamic model of the dual-arm aerial manipulator system can be derived as follows \cite{Tase}:
\begin{align}
m_t \ddot {\bm p} + m_t g \bm e_3 + \bm F_m + \bm F_f + \bm F_d = \bm U_c,
\end{align}
where
\begin{align}
\bm F_m =& \, m_t R \left[{\bm \omega} \times \left({\bm \omega} \times {\bm r}_{oc} \right)  \right] \nonumber \\
 &+ m_t R \left(\dot {\bm \omega} \times {\bm r}_{oc} + 2 \bm \omega \times \dot{\bm r}_{oc} + \ddot{\bm r}_{oc} \right),
\end{align}
and $m_t$ is the total mass of the whole system, $\bm e_3 = [0, 0, 1]^\top$ is the unit vector, and $\bm F_m \in \mathbb R^3$ denotes the force effects exerting on the multirotor caused by the manipulator, $\bm F_f(\dot{\bm p}) \in \mathbb R^3$ denotes friction, $\bm F_d \in \mathbb R^3$ denotes the general unknown nonlinear disturbance (e.g., unmodeled effects and unconsidered effects of parameters deviations), $\bm U_c = [U_{cx}, U_{cy}, U_{cz}] ^\top \in \mathbb R^3$ denotes the control input generated by the multirotor, and $\bm r_{oc}$ denotes the position of the dual-arm's CoM with respect to $\mathcal{F}_B$.

Further, the model of the system can be rewritten into the following form:
\begin{align}
\label{model1}
m_t \ddot {\bm p} + \bm F_f + \bm F = \bm U,
\end{align}
where
\begin{align}
\label{model2}
\bm U =& \, \bm U_c - \bm F_c, \\
\label{model3}
\bm F_c =& \, m_t g \bm e_3 + m_t {R} \left[{\bm \omega} \times \left({\bm \omega} \times {\bm r}_{oc} \right)  \right], \\
\label{model4}
\bm F =& \, \bm F_d + m_t {R} \left(\dot {\bm \omega} \times {\bm r}_{oc} + 2 \bm \omega \times \dot{\bm r}_{oc} + \ddot{\bm r}_{oc} \right).
\end{align}
In equations (\ref{model2})--(\ref{model4}), $\bm U = [U_x, U_y, U_z]^\top \in \mathbb{R}^3$ represents the to-be-designed virtual control input vector, $\bm F_c \in \mathbb{R}^3$ represents the force effect which can be feedforward compensated directly, and $\bm F \in \mathbb{R}^3$ represents the force effect that is unavailable due to unknown term $\bm F_d$ and high-order signals $\dot {\bm \omega}, \dot{\bm r}_{oc}, \ddot{\bm r}_{oc}$.

Considering the practical cases, the following assumption is adopted without loss of generality \cite{assu, assu2}:
\begin{assumption}
\label{ass1}
The unknown lumped uncertainty $\bm F$ and its first two time derivatives are bounded, i.e.,
\begin{align}
\Vert \bm F \Vert \leq \delta_0, \Vert \dot{\bm F} \Vert \leq \delta_1, \Vert \ddot{\bm F} \Vert \leq \delta_2,
\end{align}
where $\delta_0, \delta_1, \delta_2$ are unknown positive constants.
\end{assumption}

\section{DNN-Based Adaptive RISE Feedback Control Design} \label{sec:controller}
\begin{figure}[htbp]
  \centering
  \includegraphics[width=3.5in]{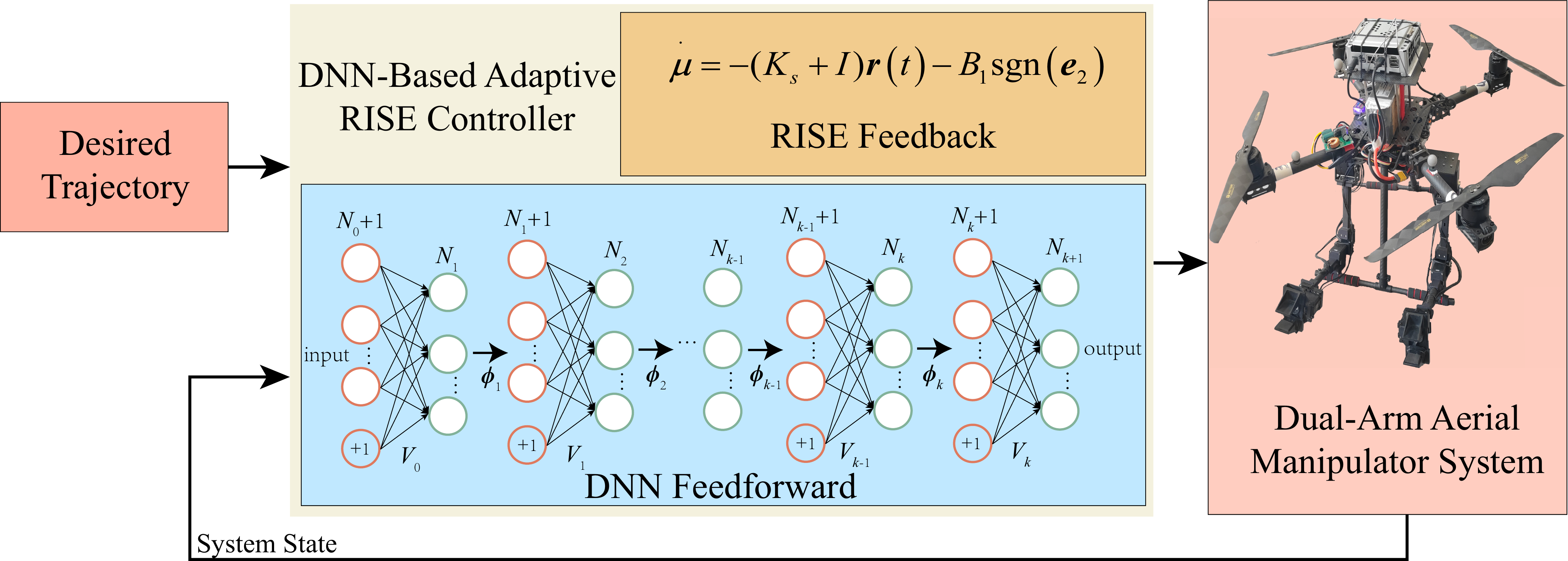}
  \caption{Structure of the control system.}
  \label{fig:DNN}
  \vspace{-0.4cm}
\end{figure}

In this section, a DNN-based adaptive RISE feedback control law is designed to drive the dual-arm unmanned aerial manipulator system to asymptotically track a desired trajectory. Fig. \ref{fig:DNN} illustrates the control structure of the system.

\subsection{Dynamics of filtered tracking error}
To quantify the control objective, the position tracking error of the multirotor, denoted by $\bm e_1 \in \mathbb{R}^3$, is firstly defined as
\begin{align}
\label{error:e1}
\bm e_1 &= \bm p - \bm p_d,
\end{align}
where $\bm p_d = [p_{xd}, p_{yd}, p_{zd}]^\top \in \mathbb R^3$ represents the multirotor's desired time-varying trajectory, and filtered tracking error signals $\bm e_2, \bm r \in \mathbb R^3$ are defined as
\begin{align}
\label{error:e2}
\bm e_2 &= \dot{\bm e}_1 + k_1 \bm e_1, \\
\label{error:r}
\bm r &= \dot{\bm e}_2 + k_2 \bm e_2,
\end{align}
where $k_1, k_2 \in \mathbb R$ are positive constants.

\begin{property}
The chosen desired trajectory $\bm p_d \in \mathbb{C}^3$, and its first, second, and third derivatives are bounded.
\end{property}

Multiplying (\ref{error:r}) by $m_t$ and combining the expressions in (\ref{model1}), (\ref{error:e1}), and (\ref{error:e2}), one can obtain the open-loop tracking error system as:
\begin{align}
\label{DNN:mr}
m_t \bm r &= m_t (k_1 \dot{\bm e}_1 + k_2 \bm e_2) + m_t (\ddot{\bm p} - \ddot{\bm p}_d) \nonumber \\
&= m_t (k_1 \dot{\bm e}_1 + k_2 \bm e_2) - m_t  \ddot{\bm p}_d - \bm F_f - \bm F + \bm U \nonumber \\
&= - m_t \ddot{\bm p}_d - \bm f_d + \bm S - \bm F + \bm U,
\end{align}
where the explicit expressions of the auxiliary functions $\bm f_d (\dot{\bm p}_d)$ and $\bm S (\bm p, \dot{\bm p}, \bm p_d, \dot{\bm p}_d)$ are as follows:
\begin{align}
&\bm f_d (\dot{\bm p}_d) = \bm F_f(\dot{\bm p}_d), \nonumber \\
&\bm S (\bm p, \dot{\bm p}, \bm p_d, \dot{\bm p}_d) = m_t (k_1 \dot{\bm e}_1 + k_2 \bm e_2) \!-\! \bm F_f(\dot{\bm p}) \!+\! \bm F_f(\dot{\bm p}_d).
\end{align}

\subsection{DNN approximation}
In this paper, a deep neural network with $k$ hidden-layers is constructed, as shown in Fig. \ref{fig:DNN}. Based on the property of universal approximation for the deep neural network \cite{DNN_univ}, the expression for $\bm f_d$ and its approximation $\hat{\bm f}_d$ can be represented as follows:
\begin{align}
\label{DNN:fd}
\bm f_d \! =& \underbrace{V_k^\top \bm\phi_k \! \left\{ V_{k-1}^\top \bm\phi_{k-1} ... \left[ V_1^\top \bm\phi_1 \left( V_0^\top \bm x_d \right) \right] \right\} }_{{\bm\Phi}_k}\! + \bm \varepsilon(\bm x_d), \\
\label{DNN:fd:hat}
\hat{\bm f}_d \! =& \underbrace{\hat V_k^\top \bm\phi_k \! \left\{ \hat V_{k-1}^\top \bm\phi_{k-1} ... \! \left[ \hat V_1^\top \bm\phi_1 \left( \hat V_0^\top \bm x_d \right) \right] \right\} \!}_{\hat{\bm\Phi}_k},
\end{align}
and the DNN architecture can also be represented recursively as
\begin{align}
\bm\Phi_j =
\left\{
\begin{aligned}
&V_j^\top \bm \phi_j (\bm \Phi_{j-1}),& j = 1, 2, ..., k, \\
&V_0^\top \bm x,& j = 0,
\end{aligned}
\right.\\
\hat{\bm\Phi}_j =
\left\{
\begin{aligned}
&\hat V_j^\top \bm \phi_j (\hat{\bm \Phi}_{j-1}),& j = 1, 2, ..., k, \\
&\hat V_0^\top \bm x,& j = 0,
\end{aligned}
\right.
\end{align}
where the input vector is chosen as $\bm x_d = \left[ \dot{\bm p}_d^\top, 1 \right]^\top \in \mathbb{R}^{N_0+1}$, $V_i, \hat V_i \in \mathbb{R}^{(N_i + 1) \times N_{i+1}}, i = 0, 1, ..., k$ denote the target constant weight matrices and the estimates, respectively, $N_0$ is the number of neurons in the input layer, $N_i$ is the number of neurons in the $i^\mathrm{th}$ hidden layer, and $N_{k+1}$ is the number of neurons in the output layer, $\bm \phi_i = \left[ \phi_{i, 1}, ..., \phi_{i, N_j}, 1 \right] \in \mathbb{R}^{(N_i + 1)}, i = 1, 2, ..., k$ denotes the activation function vector, where $\phi_{i, j}$ denotes the activation function at the $j^{\mathrm{th}}$ node of the $i^{\mathrm{th}}$ hidden layer, and $\bm \varepsilon(\bm x_d) \in \mathbb{R}^{N_{k+1}}$ denotes the functional reconstruction error, which is bounded. The target constant weight matrices are assumed to be bounded, i.e.,
\begin{align}
  \label{weight_bound}
\Vert V_i \Vert_F^2 = tr(V_i^\top V_i) \leq \bar{V}_i, i = 0, 1, ..., k,
\end{align}
where $\Vert \cdot \Vert_F$ is the Frobenius norm of a matrix, $tr(\cdot)$ is the trace of a matrix, and $\bar V_i$ is a known positive constant. Note that the input vector $\bm x_d$ and activation function vector $\bm \phi_j$ are augmented with $1$ to facilitate the inclusion of a bias term. For ease of description, the following shorthand expressions are given:
\begin{align}
\bm \phi_j = \bm \phi_j (\bm \Phi_{j-1}), \hat{\bm \phi}_j = \bm \phi_j (\hat{\bm \Phi}_{j-1}).
\end{align}
In addition, the estimation errors of the weight matrices are defined as follows:
\begin{align}
\label{DNN:Vtilde}
\tilde V_i = V_i - \hat V_i, i = 0, 1, 2, ..., k.
\end{align}
\begin{property}
Based on \textsl{Property 1}, one can conclude that $\bm \varepsilon(\bm x_d)$ and its time derivative are bounded, i.e.,
\begin{align}
\Vert \bm \varepsilon(\bm x_d) \Vert \leq \varepsilon_0, \Vert \dot{\bm \varepsilon}(\bm x_d) \Vert \leq \varepsilon_1, \Vert \ddot{\bm \varepsilon}(\bm x_d) \Vert \leq \varepsilon_2,
\end{align}
where $\varepsilon_0, \varepsilon_1, \varepsilon_2$ are known positive constants.
\end{property}

\subsection{Controller design}
Based on the result in (\ref{DNN:mr}), the nonlinear controller, including a RISE feedback control term $\bm \mu$ and a DNN feedforward control term $\hat{\bm f}_d = \left[\hat f_{dx}, \hat f_{dy}, \hat f_{dz}\right]^\top$, is designed as follows:
\begin{align}
\label{DNN:v}
\bm U =\,& \bm \mu + \hat{\bm f}_d + m_t \ddot{\bm p}_d, \\
\label{DNN:mu}
\bm \mu =\,& - (K_s + I) \bm e_2 \left( t \right) + (K_s + I) \bm e_2 \left( 0 \right) \nonumber \\
\,&- \int_0^t \left[ (K_s + I) k_2 \bm e_2 \left( \sigma \right) + B_1 \, \mathrm{sgn} \left( \bm e_2 \left( \sigma \right) \right) \right] \mathrm d\sigma,
\end{align}
where $K_s \!=\! \mathrm{diag}\! \left( \left[k_{sx},\! k_{sy},\! k_{sz} \right] \right)\!, B_1 \!=\! \mathrm{diag}\! \left( \left[\beta_{1x},\! \beta_{1y},\! \beta_{1z} \right] \right) \in \mathbb{R}_{\mathrm{+}}^{3 \times 3}$ are positive definite diagonal matrices, and $I \in \mathbb{R}^{3 \times 3}$ is the identity matrix. The control gain matrix $B_1$ is selected as follows:
\begin{align}
  \label{beta1}
  \beta_{1\mathrm{min}} \!=\! \mathrm{min}\{ \beta_{1x}, \beta_{1y}, \beta_{1z} \} \!>\! \zeta_1 + \zeta_2 + k_2^{-1} \left( \zeta_3 + \zeta_4 \right),
\end{align}
where $\zeta_1, \zeta_2, \zeta_3, \zeta_4$ are introduced in (\ref{NN:bound2}).
The time derivative of (\ref{DNN:mu}) can be calculated as
\begin{align}
\label{DNN:dmu}
\dot {\bm \mu} = - (K_s + I) \bm r \left( t \right) - B_1 \, \mathrm{sgn} \left( \bm e_2 \right).
\end{align}
The specific expression for $\hat{\bm f}_d$ is given in (\ref{DNN:fd:hat}), and the weight matrices are updated online according to the following adaption law using a continuous projection:
\begin{align}
\label{DNN:dhatW}
\dot{\hat V}_i \! = \! \mathrm{proj} \! \left( \! - \Gamma_i \hat{\bm \phi}_i^\prime \hat V_{i-1}^\top \! ... \hat{\bm \phi}_1^\prime \! \hat V_0^\top \! \dot{\bm x}_d \bm e_2^\top \! \hat V_k^\top \! \hat{\bm \phi}_k^\prime ... \! \hat V_{i+1}^\top \! \hat{\bm \phi}_{i+1}^\prime \! \right) \!,
\end{align}
where $\Gamma_i \in \mathbb{R}^{(N_i + 1)\times (N_i + 1)}, i = 0, 1, 2, ..., k$ is a positive definite diagonal gain matrix, $\hat{\bm \phi}_i^\prime \in \mathbb{R}^{(N_i + 1)\times N_i}$ is the gradient of the activation function vector at the $i^\mathrm{th}$ layer, and $\bm \phi_i^\prime (\bm a) = \frac{\partial}{\partial \bm b} \bm \phi_i(\bm b)\big\vert_{\bm b = \bm a}, \forall \bm a \in \mathbb{R}^{N_i}$.

Subsequently, taking the derivative of (\ref{DNN:fd})--(\ref{DNN:fd:hat}) with respect to time results in
\begin{align}
\label{DNN:dfd}
\dot{\bm f_d} =&\, V_k^\top \bm\phi_k^\prime V_{k-1}^\top \bm\phi_{k-1}^\prime ...  V_1^\top \bm\phi_1^\prime V_0^\top \dot{\bm x}_d + \dot{\bm \varepsilon},  \\
\label{DNN:dfd:hat}
\dot{\hat{\bm f_d}} =&\, \dot{\hat V}_k^\top \hat{\bm \phi}_k + \hat V_k^\top \hat{\bm \phi}_k^\prime \dot{\hat V}_{k-1}^\top \hat{\bm \phi}_{k-1} + ... \nonumber \\
 &+ \hat V_k^\top \hat{\bm \phi}_k^\prime {\hat V}_{k-1}^\top \hat{\bm \phi}_{k-1}^\prime ... \hat V_2^\top \hat{\bm \phi}_2^\prime \dot{\hat V}_1^\top \hat{\bm \phi}_1 \nonumber \\
 &+ \hat V_k^\top \hat{\bm \phi}_k^\prime {\hat V}_{k-1}^\top \hat{\bm \phi}_{k-1}^\prime ... \hat V_2^\top \hat{\bm \phi}_2^\prime \hat V_1^\top \hat{\bm \phi}_1^\prime \dot{\hat V}_0^\top \bm x_d \nonumber \\
 &+ \hat V_k^\top \hat{\bm \phi}_k^\prime {\hat V}_{k-1}^\top \hat{\bm \phi}_{k-1}^\prime ... \hat V_2^\top \hat{\bm \phi}_2^\prime \hat V_1^\top \hat{\bm \phi}_1^\prime \hat V_0^\top \dot{\bm x}_d.
\end{align}
Employing the results in (\ref{DNN:v}), the time derivative of (\ref{DNN:mr}) can be reduced to
\begin{align}
\label{DNN:dmr1}
m_t \dot{\bm r} =& - \dot{\bm f_d} + \dot{\hat{\bm f_d}} + \dot{\bm S} - \dot{\bm F} + \dot{\bm \mu}.
\end{align}
Define an auxiliary vector $\bm \iota = \sum_{i=0}^{k} \bm \iota_i$, where
\begin{align}
\bm \iota_i =
\left\{
\begin{aligned}
&\hat V_k^\top \hat{\bm \phi}_k^\prime ... {\tilde V}_i^\top \hat{\bm \phi}_i^\prime ... \hat V_1^\top \hat{\bm \phi}_1^\prime \hat V_0^\top \dot{\bm x}_d,& i= 0, 1, ..., k - 1, \\
& V_k^\top \hat{\bm \phi}_k^\prime {\hat V}_{k-1}^\top \hat{\bm \phi}_{k-1}^\prime ... \hat V_1^\top \hat{\bm \phi}_1^\prime \hat V_0^\top \dot{\bm x}_d,& i = k.
\end{aligned}
\right. \nonumber
\end{align}
Substituting (\ref{DNN:dmu}) and (\ref{DNN:dfd})--(\ref{DNN:dfd:hat}) into (\ref{DNN:dmr1}) leads to the following result:
\begin{align}
\label{DNN:dmr2}
m_t \dot{\bm r}
= & - V_k^\top \bm\phi_k^\prime V_{k-1}^\top \bm\phi_{k-1}^\prime ...  V_1^\top \bm\phi_1^\prime V_0^\top \dot{\bm x}_d - \dot{\bm \varepsilon} + \dot{\hat
    V}_k^\top \hat{\bm \phi}_k \nonumber \\
  & + \hat V_k^\top \hat{\bm \phi}_k^\prime \dot{\hat V}_{k-1}^\top \hat{\bm \phi}_{k-1} + ... \nonumber \\
  & + \hat V_k^\top \hat{\bm \phi}_k^\prime {\hat V}_{k-1}^\top \hat{\bm \phi}_{k-1}^\prime ... \hat V_2^\top \hat{\bm \phi}_2^\prime \dot{\hat V}_1^\top \hat{\bm \phi}_1 \nonumber \\
  & + \hat V_k^\top \hat{\bm \phi}_k^\prime {\hat V}_{k-1}^\top \hat{\bm \phi}_{k-1}^\prime ... \hat V_2^\top \hat{\bm \phi}_2^\prime \hat V_1^\top \hat{\bm \phi}_1^\prime \dot{\hat V}_0^\top \bm x_d \nonumber \\
  & + \hat V_k^\top \hat{\bm \phi}_k^\prime {\hat V}_{k-1}^\top \hat{\bm \phi}_{k-1}^\prime ... \hat V_2^\top \hat{\bm \phi}_2^\prime \hat V_1^\top \hat{\bm \phi}_1^\prime \hat V_0^\top \dot{\bm x}_d \nonumber \\
  & + \dot{\bm S} - \dot{\bm F} - (K_s + I) \bm r \left( t \right) - B_1 \, \mathrm{sgn} \left( \bm e_2 \right).
\end{align}
After adding and subtracting the term $\bm \iota$ to (\ref{DNN:dmr2}), and with some further arrangements, one can obtain the following conclusion:
\begin{align}
\label{DNN:dmr3}
m_t \dot{\bm r} \!=\! & - \!(K_s \!+\! I) \bm r \left( t \right) \!-\! B_1 \mathrm{sgn} \left( \bm e_2 \right) - \bm e_2 + \tilde{\bm N} + \bm N,
\end{align}
where $\bm N = \bm N_d + \bm N_{\eta}, \bm N_\eta = \bm N_{\eta_1} + \bm N_{\eta_2}$, and
\begin{align}
\tilde{\bm N} =&\, \dot{\hat V}_k^\top \hat{\bm \phi}_k + \hat V_k^\top \hat{\bm \phi}_k^\prime \dot{\hat V}_{k-1}^\top \hat{\bm \phi}_{k-1} + ... + \dot{\bm S} + \bm e_2 \nonumber \\
 &+ \hat V_k^\top \hat{\bm \phi}_k^\prime {\hat V}_{k-1}^\top \hat{\bm \phi}_{k-1}^\prime ... \hat V_2^\top \hat{\bm \phi}_2^\prime \dot{\hat V}_1^\top \hat{\bm \phi}_1 \nonumber \\
 &+ \hat V_k^\top \hat{\bm \phi}_k^\prime {\hat V}_{k-1}^\top \hat{\bm \phi}_{k-1}^\prime ... \hat V_2^\top \hat{\bm \phi}_2^\prime \hat V_1^\top \hat{\bm \phi}_1^\prime \dot{\hat V}_0^\top \bm x_d,  \\
\bm N_d =& - V_k^\top \bm\phi_k^\prime V_{k-1}^\top \bm\phi_{k-1}^\prime ...  V_1^\top \bm\phi_1^\prime V_0^\top \dot{\bm x}_d - \dot{\bm \varepsilon} - \dot{\bm F}, \\
\bm N_{\eta_1} =&\, \bm{\iota}_0 + \bm{\iota}_1 + ... + \bm{\iota}_{k-1} + \bm{\iota}_k, \\
\bm N_{\eta_2} =& - \bm{\iota}_0 - \bm{\iota}_1 - ... - \bm{\iota}_{k-1} \nonumber \\
  &- \tilde V_k^\top \hat{\bm \phi}_k^\prime \hat V_{k-1}^\top \hat{\bm \phi}_{k-1}^\prime ... \hat V_1^\top \hat{\bm \phi}_1^\prime \hat V_0^\top \dot{\bm x}_d.
\end{align}
Based on the adaption law (\ref{DNN:dhatW}), the Mean Value Theorem can be utilized to derive the following upper bound:
\begin{align}
\label{NN:bound1}
\Vert \tilde{\bm N}(t) \Vert \leq \rho \left( \Vert \bm z \Vert \right) \Vert \bm z \Vert,
\end{align}
where $\bm z(t) = \left[ \bm e_1^\top, \bm e_2^\top,  \bm r^\top \right]^\top \in \mathbb{R}^{9}$, and the bounding function $\rho \left( \Vert \bm z \Vert \right) \in \mathbb R$ is a positive globally invertible nondecreasing function. In addition, the following inequalities can be developed based on \textsl{Assumption 1}, \textsl{Property 2} and (\ref{weight_bound}):
\begin{align}
\label{NN:bound2}
\Vert \bm N_d \Vert &\leq \zeta_1, \Vert \bm N_\eta \Vert \leq \zeta_2, \nonumber \\
\Vert \dot{\bm N}_d \Vert &\leq \zeta_3, \Vert \dot{\bm N}_\eta \Vert \leq \zeta_4 + \zeta_5 \Vert \bm e_2 \Vert,
\end{align}
where $\zeta_i \in \mathbb R, i = 1, 2, ..., 5$ are positive constants.

\section{Stability Analysis} \label{sec:stability}
\begin{theorem}
Considering the nonlinear dual-arm unmanned aerial manipulator system, by utilizing the proposed controller (\ref{DNN:v}) and the adaptive update law (\ref{DNN:dhatW}), it can be guaranteed that the tracking error signal $\bm z = \left[ \bm e_1^\top, \bm e_2^\top,  \bm r^\top \right]^\top \in \mathbb{R}^{9}$ converges to the origin asymptotically in the sense that
\begin{align}
\lim_{t \to \infty} \bm z = \bm 0_{9 \times 1},
\end{align}
under the condition that $K_s$ is large enough and (\ref{beta1}) is satisfied.
\end{theorem}

\begin{IEEEproof}
Firstly, define the following auxiliary functions:
\begin{align}
\label{proof:L}
L =&\, \bm r^\top \! \left[ \bm N_{\eta_1} \!+\! \bm N_d \!-\! B_1 \mathrm{sgn} \! \left( \bm e_2 \right) \right] \!+\! \dot{\bm e}_2^\top \! \bm N_{\eta_2} \!-\! \bm e_2^\top \! B_2 \bm e_2, \\
\zeta_b =&\, \bm \beta_1^\top \vert \bm e_2 \left(0 \right) \vert - \bm e_2^\top \left(0 \right) \bm N \left( 0 \right), 
\end{align}
\begin{align}
P =&\, \zeta_b - \int_0^t L(\tau) \mathrm d \tau,
\end{align}
where $\bm \beta_1 \!=\! \left[\beta_{1x},\! \beta_{1y},\! \beta_{1z} \right]^\top \in \mathbb{R}^3$ is the vectorization of $B_1$, $B_2 = \mathrm{diag} \left( \left[\beta_{2x},\, \beta_{2y},\, \beta_{2z} \right] \right)\in \mathbb{R}_{\mathrm{+}}^{3 \times 3}$ is an auxiliary positive definite diagonal matrix, which satisfies that $\beta_{2\mathrm{min}} = \mathrm{min}\{ \beta_{2x},$  $\beta_{2y}, \beta_{2z} \} > \zeta_5,$ and $\vert \bm a \vert = \left[ \vert a_1 \vert, \vert a_2 \vert, \vert a_3 \vert \right]^\top$, for any vector $\bm a = \left[ a_1, a_2, a_3 \right]^\top$.

Then, based on the fact that $\frac{\mathrm d \vert x(t) \vert}{\mathrm d t} = \dot x \, \mathrm{sgn}(x)$, it can be determined that
\begin{align}
&\int_0^t L(\tau) \mathrm d\tau \nonumber \\
=& \int_0^t \left( \dot{\bm e}_2 + k_2 \bm e_2 \right)^\top \left[ \bm N_{\eta_1} + \bm N_d - B_1 \, \mathrm{sgn} \left( \bm e_2 \right) \right] \mathrm d\tau
   \nonumber \\
 & + \int_0^t \left( \dot{\bm e}_2^\top \bm N_{\eta_2} - \bm e_2^\top B_2 \bm e_2 \right) \mathrm d\tau \nonumber \\
=& \,\bm \beta_1^\top \vert \bm e_2(0) \vert - \bm e_2^\top(0) \bm N(0) + {\bm e}_2^\top \bm N - \bm \beta_1^\top \vert \bm e_2 \vert \nonumber \\
 & + \int_0^t \bm e_2^\top k_2 \left[ \bm N_{\eta_1} + \bm N_d - B_1 \, \mathrm{sgn} \left( \bm e_2 \right) \right] \mathrm d\tau \nonumber \\
 & - \int_0^t {\bm e}_2^\top \left( \dot{\bm N}_{d} + \dot{\bm N}_\eta \right) \mathrm d\tau - \int_0^t \bm e_2^\top B_2 \bm e_2 \mathrm d\tau,
\end{align}
combining with (\ref{NN:bound1}) and (\ref{NN:bound2}), the following conclusion can be deduced:
\begin{align}
\label{DNN:L}
&\int_0^t L(\tau) \mathrm d\tau \nonumber \\
\leq& \,\bm \beta_1^\top \vert \bm e_2(0) \vert - \bm e_2^\top(0) \bm N(0) + \Vert {\bm e}_2 \Vert \Vert \bm N \Vert - \beta_{1\mathrm{min}} \Vert \bm e_2 \Vert \nonumber \\
& + k_2 \int_0^t \Vert \bm e_2^\top \Vert \left(\Vert \bm N_{\eta} \Vert + \Vert \bm N_d \Vert - \beta_{1\mathrm{min}} \right) \mathrm d\tau \nonumber \\
& + k_2 \int_0^t \Vert \bm e_2^\top \Vert \left( k_2^{-1} \Vert \dot{\bm N}_d \Vert  + k_2^{-1} \Vert \dot{\bm N}_\eta \Vert \right) \mathrm d\tau \nonumber \\
& - \int_0^t \beta_{2\mathrm{min}} \Vert \bm e_2 \Vert^2 \mathrm d\tau \nonumber \\
=& \, \bm \beta_1^\top \vert \bm e_2(0) \vert - \bm e_2^\top(0) \bm N(0) + \Vert {\bm e}_2 \Vert \left( \Vert \bm N \Vert - \beta_{1\mathrm{min}} \right) \nonumber \\
& + k_2 \int_0^t \Vert \bm e_2^\top \Vert \left(\zeta_1 + \zeta_2 + k_2^{-1} \left( \zeta_3 + \zeta_4 \right) - \beta_{1\mathrm{min}} \right) \mathrm d\tau \nonumber \\
& + \int_0^t \left( \zeta_5 - \beta_{2\mathrm{min}} \right) \Vert \bm e_2 \Vert^2 \mathrm d\tau.
\end{align}

It is obvious that the following results can be obtained:
\begin{align}
\int_0^t L(\tau) d\tau \leq \zeta_b,
\end{align}
that is to say, $P \geq 0$. Taking the time derivative of $P$, one can derive
\begin{align}
\dot P \!=&\! - \! \bm r^\top \! \left[ \bm N_{\eta_1} \!+\! \bm N_d \!-\! B_1 \mathrm{sgn} \! \left( \bm e_2 \right) \right] \!-\! \dot{\bm e}_2^\top \! \bm N_{\eta_2} \!+\! \bm e_2^\top \! B_2 \bm e_2.
\end{align}
Based on the above analysis, choosing the Lyapunov function candidate as follows:
\begin{align}
\label{NN:V}
E = \bm e_1^\top \bm e_1 + \frac{1}{2} \bm e_2^\top \bm e_2 + \frac{1}{2} \bm r^\top \bm r + P + Q,
\end{align}
where $Q$ is a quadratic positive definite function:
\begin{align}
Q = \frac{k_2}{2} \sum_{i=0}^k tr \left( \tilde V_i^\top \Gamma_i^{-1} {\tilde V}_i \right).
\end{align}
The time derivative of $E$ can be calculated as:
\begin{align}
\label{DNN:dV}
\dot E =&\, 2\bm e_1^\top \dot{\bm e}_1 + \bm e_2^\top \dot{\bm e}_2 + \bm r^\top \dot{\bm r} + \dot P + \dot Q \nonumber \\
=&\, 2 \bm e_1^\top \bm e_2 - 2 \bm e_1^\top k_1 \bm e_1 - \bm e_2^\top k_2 \bm e_2 + \bm r^\top \tilde{\bm N} - \bm r^\top (K_s \nonumber 
\end{align}
\begin{align}
 & + I) \bm r + \bm e_2^\top B_2 \bm e_2 + k_2 \sum_{i=0}^k tr \left( \tilde V_i^\top \Gamma_i^{-1} \dot{\tilde V}_i \right) \vspace{0.0ex} \nonumber \\
 & + \! \bm e_2^\top \! k_2 \! \left( \sum_{i=0}^k \! \hat V_k^\top \! \hat{\bm \phi}_k^\prime \! \hat V_{k-1}^\top \! \hat{\bm \phi}_{k-1}^\prime \! ... \tilde V_i^\top \! \hat{\bm \phi}_i^\prime \! ... \hat V_0^\top \! \hat{\bm \phi}_0^\prime \dot{\bm x}_d \! \right).
\end{align}
Combining with (\ref{DNN:Vtilde}) and substituting (\ref{DNN:dhatW}) into (\ref{DNN:dV}) yields the following result:
\begin{align}
\dot E =&\, 2 \bm e_1^\top \bm e_2 - 2 \bm e_1^\top k_1 \bm e_1 - \bm e_2^\top k_2 \bm e_2 + \bm r^\top \tilde{\bm N}  \nonumber \\
 & - \bm r^\top (K_s + I) \bm r + \bm e_2^\top B_2 \bm e_2 \nonumber \\
 \leq& - (2 k_1 - 1) \Vert \bm e_1 \Vert^2 - (k_2 - \beta_{2\mathrm{max}} - 1) \Vert \bm e_2 \Vert^2 \nonumber \\
 & -  \Vert \bm r \Vert^2 - \left( k_{s\mathrm{min}} - \rho(\Vert \bm z \Vert) \Vert \bm r \Vert \Vert \bm z \Vert \right) \nonumber \\
 \leq& - \lambda \Vert \bm z \Vert^2 - \left( k_{s\mathrm{min}} - \rho(\Vert \bm z \Vert) \Vert \bm r \Vert \Vert \bm z \Vert \right) \nonumber \\
 =& - \left( \lambda - \frac{\rho^2(\Vert \bm z \Vert)}{4 k_{s\mathrm{min}}} \right) \Vert \bm z \Vert^2,
\end{align}
where $\beta_{2\mathrm{max}} = \mathrm{max}\{ \beta_{2x},$  $\beta_{2y}, \beta_{2z} \}, k_{s\mathrm{min}} = \mathrm{min}\{ k_{sx}, k_{sy}, k_{sz} \}, \lambda = \mathrm{min}\{ 2k_1 - 1, k_2 - \beta_{2\mathrm{max}} - 1, 1 \}$. $\lambda$ is positive if $k_1, k_2$ satisfy the following sufficient conditions:
\begin{align}
k_1 > \frac{1}{2}, k_2 > \beta_{2\mathrm{max}} + 1.
\end{align}
Choosing $K_s$ large enough according to the following condition:
\begin{align}
k_{s\mathrm{min}} \geq \frac{\rho^2(\Vert \bm z \Vert)}{\lambda},
\end{align}
further yields $\dot E \leq 0$, implying that the tracking error signal $\bm z$ converges to zero asymptotically, i.e., the proof of \textsl{Theorem 1} is accomplished.

\begin{figure}[t]
  \centering
  \includegraphics[width=3.0in]{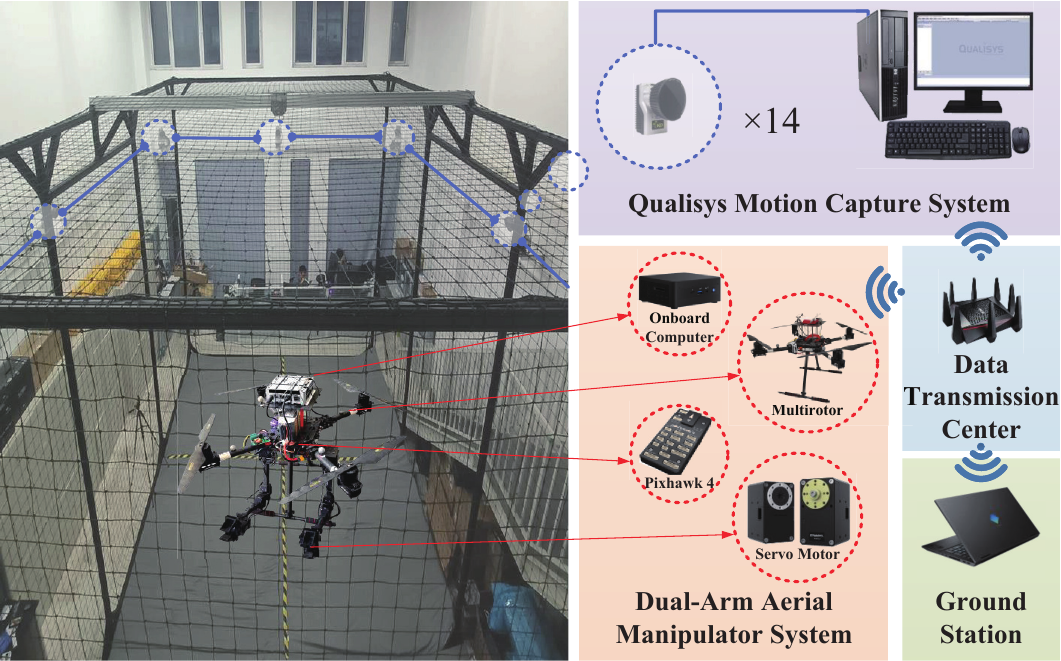}
  \caption{Hardware experimental platform.}
  \label{fig:tb}
  \vspace{-0.4cm}
\end{figure}
  
\begin{figure}[t]
  \centering
  \includegraphics[width=3.2in]{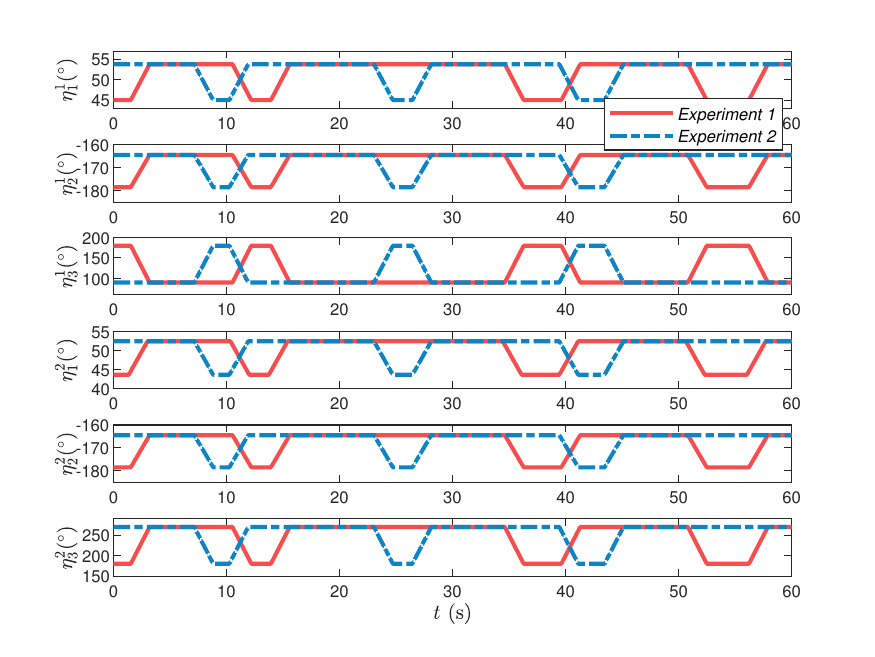}
  \caption{Joint angles of the dual-arm.}
  \label{fig:sroc}
  \vspace{-0.6cm}
\end{figure}

\begin{figure*}[t]
  \centering
  \clearcaptionsetup{figure}
  \clearcaptionsetup{subfloat}
  \captionsetup[subfloat]{labelsep=period}
  \subfloat[\small \emph{Experiment 1.}]{\includegraphics[width=0.3\textwidth]{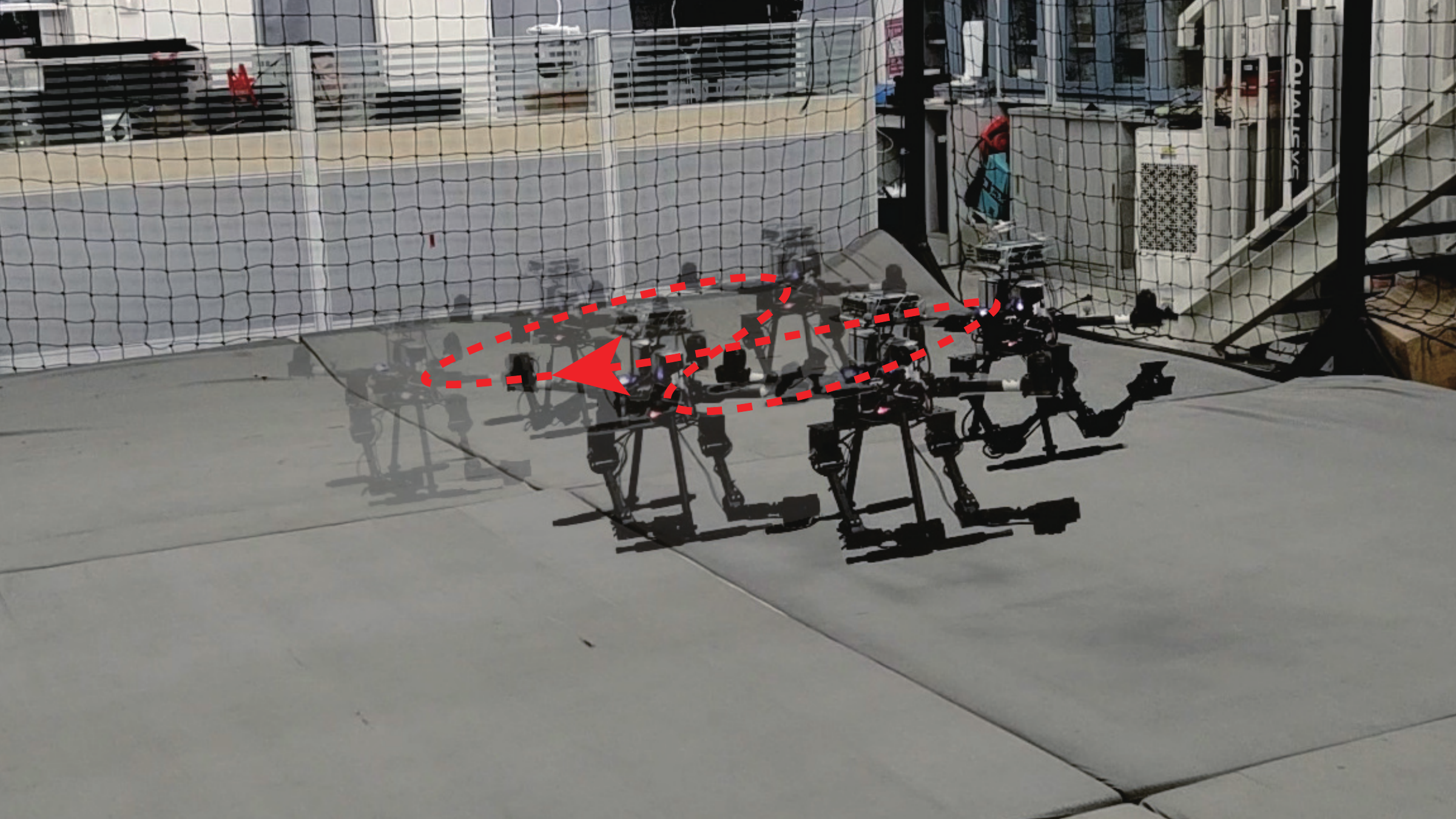}}\hspace{0.0mm}
  \label{exp_pica}
  \captionsetup[subfloat]{labelsep=period}
  \subfloat[\small \emph{Experiment 2.}]{\includegraphics[width=0.3\textwidth]{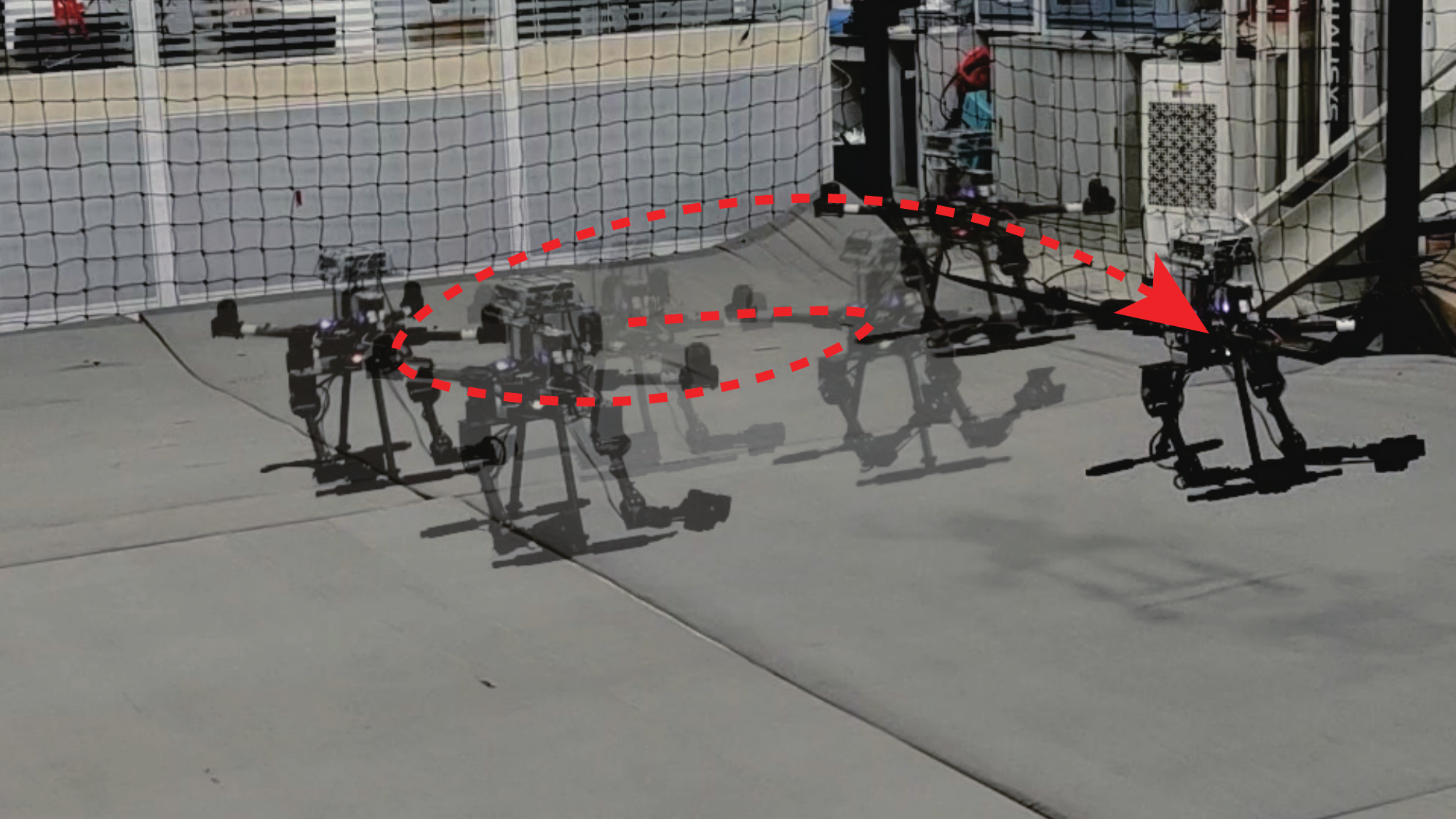}}\hspace{0.0mm}
  \label{exp_picb}
  \captionsetup[subfloat]{labelsep=period}
  \subfloat[\small \emph{Experiment 3.}]{\includegraphics[width=0.3\textwidth]{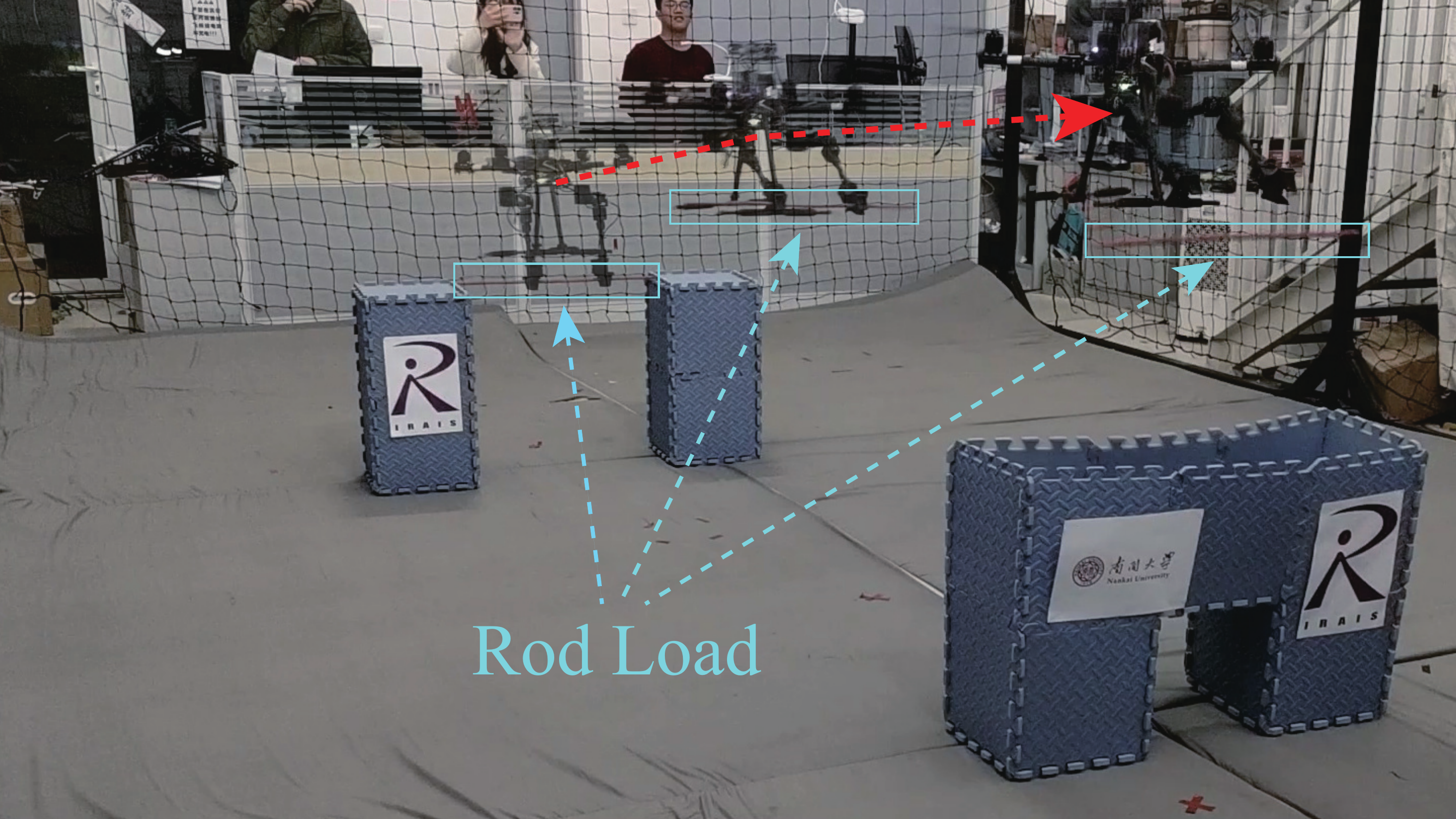}}\hspace{0.0mm}
  \label{exp_picc}
  \caption{Snaps of the three groups of actual experiments. }
  \label{exp_pic}
  \vspace{-0.3cm}
\end{figure*}

\begin{figure*}[t]
  \centering
  \clearcaptionsetup{figure}
  \clearcaptionsetup{subfloat}
  \captionsetup[subfloat]{labelsep=period}
  \subfloat[\small Multirotor position.]{\includegraphics[width=0.32\textwidth]{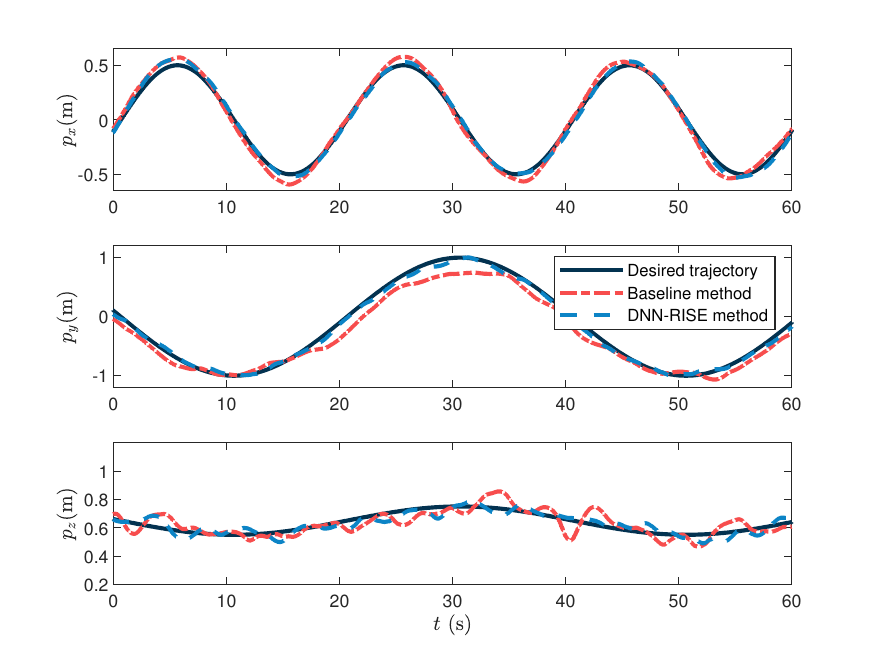}}\hspace{0.0mm}
  \captionsetup[subfloat]{labelsep=period}
  \subfloat[\small Force inputs.]{\includegraphics[width=0.32\textwidth]{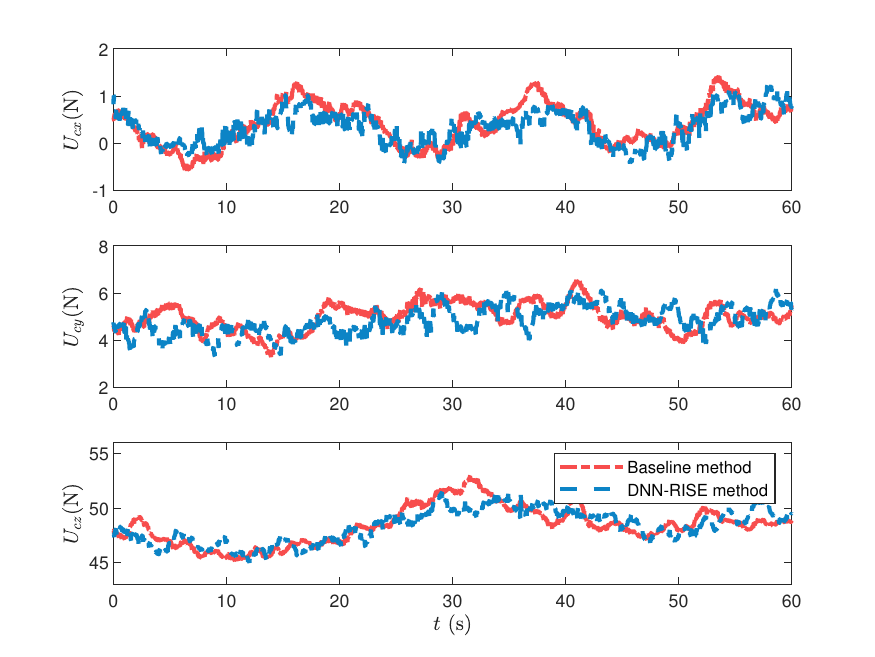}}\hspace{0.0mm}
  \captionsetup[subfloat]{labelsep=period}
  \subfloat[\small DNN outputs.]{\includegraphics[width=0.32\textwidth]{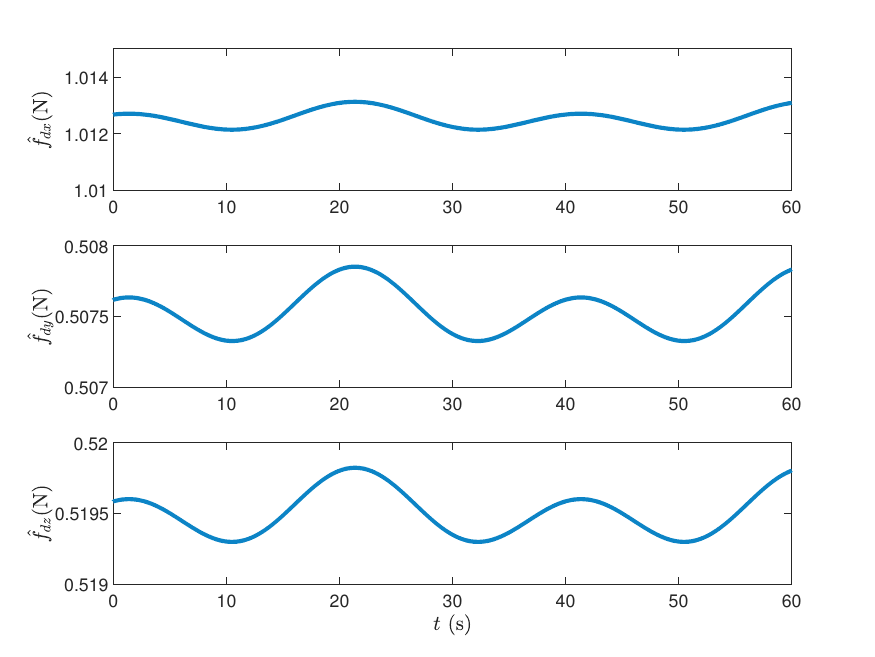}}\hspace{0.0mm}
  \caption{Results of \emph{Experiment 1}.}
  \label{fig:exp1}
\end{figure*}

\end{IEEEproof}

\section{Hardware Experimental Validation}\label{sec:exp}

\begin{table}[b]\scriptsize
  \centering
  \captionsetup{font={small}}
  \caption{Servo Motor Parameters}
  \renewcommand\arraystretch{1.65}
  \begin{tabular}{ccccc}
  \hline
  Joint & Servo model & Torque(N$\cdot$m) & Weight(g)     \\
  \midrule
  Pitch of shoulder & XM540-W270 & 12.9 & 165 \\
  Pitch of elbow & XM430-W350 & 4.8 & 82 \\
  Roll of elbow & XM430-W350 & 4.8 & 82 \\
  Claw & AX-12A & 1.5 & 54.6 \\
  \bottomrule
      \end{tabular}
      \label{tab:para}
  \end{table}

In this section, to verify the feasibility and robustness of the proposed control scheme, three groups of experiments have been conducted on the self-built hardware platform. It is worth mentioning that the collaborative grasping and delivering experiment of a rod load is tested.

\subsection{Experimental Platform}

The self-built hardware experimental platform of the dual-arm aerial manipulator system is depicted in Fig. \ref{fig:tb}, which is comprised of a multirotor equipped with an onboard computer and the dual-arm manipulator. Data from 14 Qualisys cameras, including multirotor positions, linear velocities, and yaw angles, are transmitted to the ground station via the ROS (Robot Operating System)-based transmission protocol over the LAN (Local Area Network). A PixHawk flight control unit is linked to the onboard computer through the MAVROS communication protocol. The onboard computer NUC and ground station run the 64-bit Ubuntu 20.04 and 64-bit Ubuntu 18.04 operating systems, respectively. Control inputs are computed on the ground station and transmitted to the onboard computer via WIFI using the 5G band. For the dual-arm aerial manipulator, Dynamixel's AX and XM series are selected, with their parameters summarized in Table \ref{tab:para}. The dual-arm manipulator is under the control of the onboard computer. The physical parameters of the experimental platform are given as follows:
\begin{align}
m_t&=4.85~\mathrm{kg}, g=9.8~\mathrm{m/s^2},  \nonumber \\
L_1 &= 0.15~\mathrm{m}, L_2 = 0.05~\mathrm{m}, L_3 = 0.17~\mathrm{m}. \nonumber 
\end{align}
In practical test, the control gains are set as follows:
\begin{align}
K_s &= \mathrm{diag} \left( \left[10.0, 10.0, 14.5 \right] \right), k_1 = 0.69, \nonumber \\
B_1 &= \mathrm{diag} \left( \left[4.0, 4.0, 4.0 \right]\right), k_2 = 0.5. \nonumber
\end{align}
The DNN comprises four layers ($k = 3$): the first layer contains 3 neurons, while the last three layers each contain 4 neurons, i.e., $N_0=3, N_i = 4, i = 1, 2, 3$. The chosen activation function is the sigmoid function, and the update matrices are defined as follows:
\begin{align}
\Gamma_0 &= \mathrm{diag} \left( \left[2, 2, 2, 2 \right] \right) \times 10^6, \nonumber \\
\Gamma_i &= \mathrm{diag} \left( \left[2, 2, 2, 2, 2 \right] \right) \times 10^6, i = 1, 2, 3. \nonumber 
\end{align}
The baseline controller $\bm u = - K_p \bm e_1 - K_d \dot{\bm e}_1 + \bm F_c + m_t \ddot{\bm p}_d$ is chosen as the comparison method and the control gains are provided as follows:
\begin{align}
K_p &= \mathrm{diag} \left( \left[8.0, 8.0, 10.0 \right] \right), K_d = \mathrm{diag} \left( \left[10.0, 10.0, 13.0 \right] \right). \nonumber 
\end{align}

In the first two experiments, the multirotor follows a figure-eight and a spiral trajectory, respectively, while the dual-arm manipulator moves periodically. The different desired trajectories of the multirotor are designed to verify the robustness of the proposed method. In the third experiment, a functional verification experiment is presented where the dual-arm manipulator collaboratively grasps a rod-shaped load and delivers it to a desired position. Subsequently, the trajectory tracking tests and collaborative grasping test are conducted on the experimental platform.

\begin{figure*}[t]
  \centering
  \clearcaptionsetup{figure}
  \clearcaptionsetup{subfloat}
  \captionsetup[subfloat]{labelsep=period}
  \subfloat[\small Multirotor position.]{\includegraphics[width=0.32\textwidth]{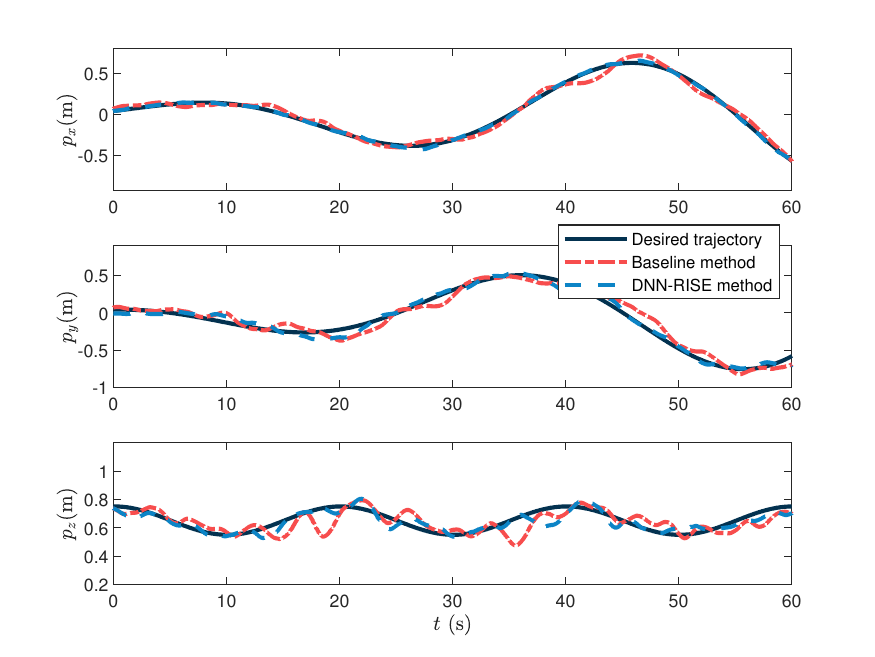}}\hspace{0.0mm}
  \captionsetup[subfloat]{labelsep=period}
  \subfloat[\small Force inputs.]{\includegraphics[width=0.32\textwidth]{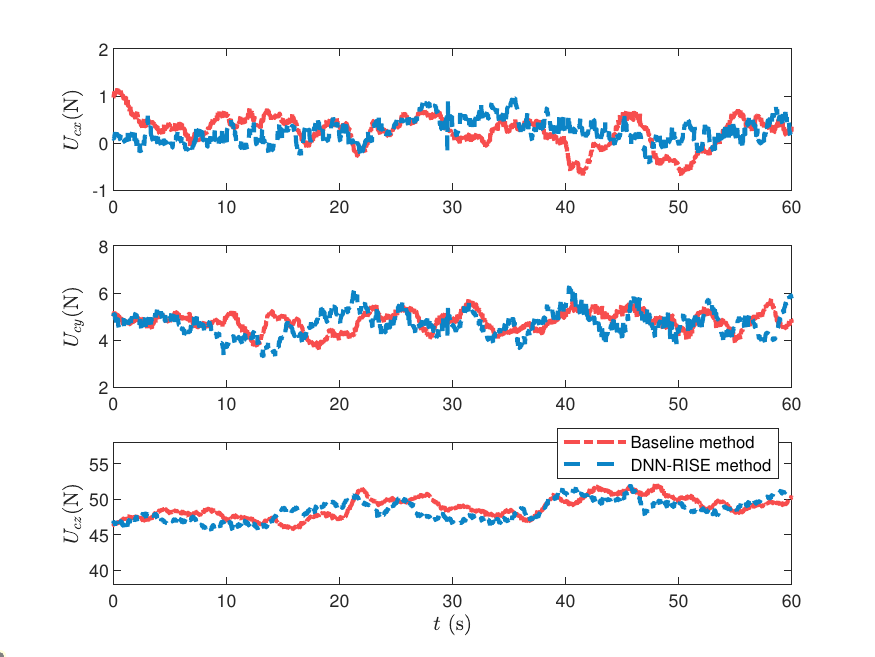}}\hspace{0.0mm}
  \captionsetup[subfloat]{labelsep=period}
  \subfloat[\small DNN outputs.]{\includegraphics[width=0.32\textwidth]{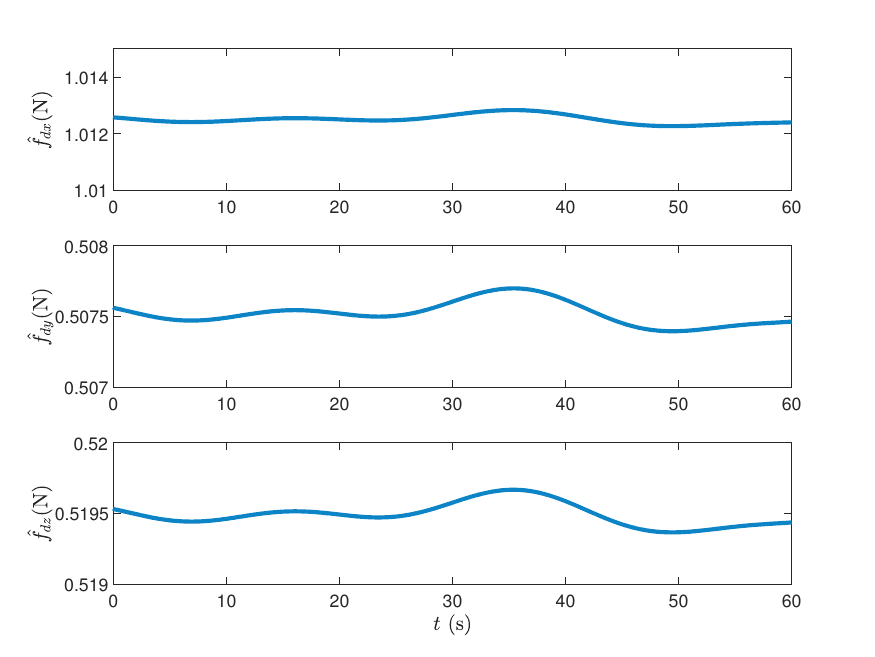}}\hspace{0.0mm}
  \caption{Results of \emph{Experiment 2}.}
  \label{fig:exp2}
  \vspace{-0.2cm}
\end{figure*}

\subsection{Experimental Results}
\subsubsection{Experiment 1 - Figure-Eight Trajectory Tracking}
This group of experiment primarily assesses the performance of the proposed algorithm under no-load conditions, which is fundamental for assembly and transportation tasks. The control objective is to guide the multirotor along a figure-eight trajectory, and the explicit expressions are as follows:
\begin{align}
  \left\{
    \begin{aligned}
      p_{xd} &= 0.5 \sin(\frac{\pi}{10}t), \nonumber \\
      p_{yd} &= \sin(\frac{\pi}{20}t + \pi), \nonumber \\
      p_{zd} &= 0.65 + 0.1 \sin(\frac{\pi}{20}t + \pi). \nonumber 
    \end{aligned}
  \right.
\end{align}
Concurrently, the joint angles of the dual-arm manipulator are set to follow the trajectories illustrated in Fig. \ref{fig:sroc}. Fig. \ref{exp_pic}a depicts the setup for \emph{Experiment 1}, with corresponding results presented in Fig. \ref{fig:exp1}. Specifically, Fig. \ref{fig:exp1}a illustrates the position of the multirotor, while Fig. \ref{fig:exp1}b and Fig. \ref{fig:exp1}c show the force inputs and the output of the DNN, respectively. The maximum and mean errors
for the multirotor's positions, as well as the percentages of error reduction of the proposed DNN-RISE method compared to the baseline method in three directions,
are summarized in Table \ref{tab:error}. It is evident that both control methods can stabilize the system near the desired trajectory, while the proposed method exhibiting better trajectory tracking accuracy. Furthermore, the proposed scheme adeptly handles the inherent challenges posed by unignorable nonlinearities and the complex dynamic coupling between the multirotor and the manipulator. In essence, the proposed method demonstrates greater effectiveness in mitigating the influence of the manipulator's motion, parameter uncertainties, and external disturbances.

\begin{table}[t]\scriptsize
  \centering
  \captionsetup{font={small}}
  \caption{Position Errors of \emph{Experiment 1}}
  \renewcommand\arraystretch{1.45}
  \begin{tabular}{c|c|c|c|c}
  \hline \hline
  Error & Method & $x~(\mathrm{m})$ & $y~(\mathrm{m})$ & $z~(\mathrm{m})$     \\ \hline
  \multirow{3}{*}{Max} & DNN-RISE & $\mathbf{0.0703}$ & $\mathbf{0.1028}$ & $\mathbf{0.0811}$ \\ \cline{2-5}
  & Baseline & $0.1193$ & $0.2818$ & $0.1445$ \\ \cline{2-5}
  & Reduced & $\mathbf{41.07\%}$ & $\mathbf{63.52\%}$ & $\mathbf{43.89\%}$ \\ \hline
  \multirow{3}{*}{Mean} & DNN-RISE & $\mathbf{0.0237}$ & $\mathbf{0.0379}$ & $\mathbf{0.0291}$ \\ \cline{2-5}
  & Baseline & $0.0425$ & $0.1284$ & $0.0422$ \\ \cline{2-5}
  & Reduced & $\mathbf{44.24\%}$ & $\mathbf{70.46\%}$ & $\mathbf{30.95\%}$ \\ \hline
      \end{tabular}
      \label{tab:error}
\end{table}

\begin{table}[t]\scriptsize
  \centering
  \captionsetup{font={small}}
  \caption{Position Errors of \emph{Experiment 2}}
  \renewcommand\arraystretch{1.45}
  \begin{tabular}{c|c|c|c|c}
  \hline \hline
  Error & Method & $x~(\mathrm{m})$ & $y~(\mathrm{m})$ & $z~(\mathrm{m})$     \\ \hline
  \multirow{3}{*}{Max} & DNN-RISE & $\mathbf{0.0467}$ & $\mathbf{0.1215}$ & $\mathbf{0.1243}$ \\ \cline{2-5}
  & Baseline & $0.1278$ & $0.1915$ & $0.2020$ \\ \cline{2-5}
  & Reduced & $\mathbf{63.45\%}$ & $\mathbf{36.56\%}$ & $\mathbf{38.46\%}$ \\ \hline
  \multirow{3}{*}{Mean} & DNN-RISE & $\mathbf{0.0091}$ & $\mathbf{0.0424}$ & $\mathbf{0.0309}$ \\ \cline{2-5}
  & Baseline & $0.0393$ & $0.0628$ & $0.0484$ \\ \cline{2-5}
  & Reduced & $\mathbf{76.81\%}$ & $\mathbf{32.47\%}$ & $\mathbf{36.02\%}$ \\ \hline
      \end{tabular}
      \label{tab:error2}
\end{table}

\begin{figure}[htbp]
  \centering
  \clearcaptionsetup{figure}
  \clearcaptionsetup{subfloat}
  \captionsetup[subfloat]{labelsep=period}
  \subfloat[\small Multirotor position.]{\includegraphics[width=0.38\textwidth]{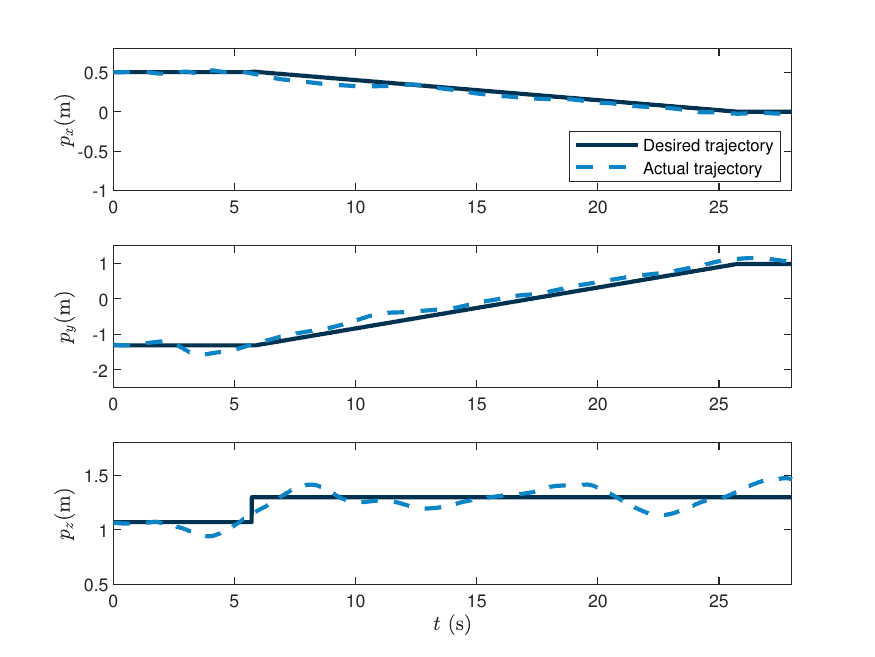}}\hspace{0.0mm}
  \captionsetup[subfloat]{labelsep=period}
  \subfloat[\small Force inputs.]{\includegraphics[width=0.38\textwidth]{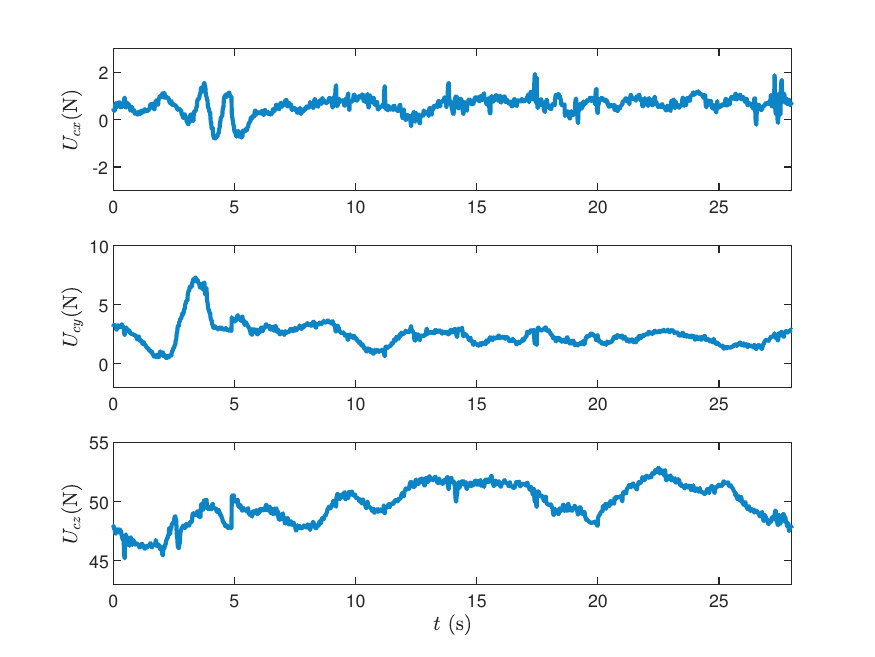}}\hspace{0.0mm}
  \caption{Results of \emph{Experiment 3}.}
  \label{fig:exp3}
  \vspace{-0.2cm}
\end{figure}

\subsubsection{Experiment 2 - Spiral Trajectory Tracking}
This set of experiment aimed at verifying the robustness of the proposed method for tracking different desired trajectories, the multirotor follows the specified spiral trajectory:
\begin{align}
  \left\{
    \begin{aligned}
      p_{xd} &= \frac{t+5}{80} \sin(\frac{\pi}{20}t+\frac{\pi}{4}), \nonumber \\
      p_{yd} &= \frac{t+5}{80} \cos(\frac{\pi}{20}t+\frac{\pi}{4}), \nonumber \\
      p_{zd} &= 0.65 + 0.1 \sin(\frac{\pi}{10}t+\frac{\pi}{2}). \nonumber 
    \end{aligned}
  \right.
\end{align}
Similar to \emph{Experiment 1}, while the multirotor tracks the spiral trajectory, the dual-arm manipulator also follows a predefined trajectory, as depicted by the curves in Fig. \ref{fig:sroc}. Fig. \ref{exp_pic}b illustrates the key snapshots of the \emph{Experiment 2}. The experimental results are presented in Fig. \ref{fig:exp2} and Table \ref{tab:error2}, showing the superior tracking performance of the proposed control scheme. It is evident that even when the multirotor tracks a different trajectory while the dual-arm manipulator is in motion, the proposed method effectively mitigates the effects caused by the manipulator's movement, showing satisfactory robustness.

\subsubsection{Experiment 3 - Collaborative Grasping and Delivery}

To explore the collaborative capabilities of the proposed scheme in achieving operational tasks, this group of experiment introduces a challenging collaborative grasping and delivery task. The dual-arm unmanned aerial manipulator system first picks up a rod load precisely from the starting point, then, transports the load to a designated location. Throughout this process, the dual-arm manipulator adjusts its joint angles to meet the task's requirements. Specifically, Fig. \ref{exp_pic}c illustrates the crucial stages of the entire process, with the results depicted in Fig. \ref{fig:exp3}. At $t = 3 \mathrm{s}$, the system reaches the starting point and successfully collaboratively grasps the rod load. At $t = 15 \mathrm{s}$, the system moves to the target location with the rod load and subsequently delivers it. Moreover, this experiment demonstrates that the dual-arm aerial manipulator system effectively exhibits cooperative operational skills in executing the designated tasks under the control of the proposed method.

\section{Conclusions}\label{sec:con}
In conclusion, this study addresses the dynamic challenges inherent in controlling a dual-arm aerial manipulator system, which has garnered significant attention from researchers. By proposing a nonlinear adaptive RISE controller incorporating DNN techiniques, this paper effectively mitigates issues arising from changing center of mass and uncertainties, enhancing control performance and ensuring operational safety. By utilizing Lyapunov techniques, the asymptotic convergence of tracking error signals is guaranteed. Real-world experiments validate the practicality and robustness of the proposed control law, providing valuable insights into its performance in handling complex scenarios. In the future, we will explore the integration of vision-based perception to further enhance the system's autonomy and adaptability, thereby expanding the scope of applications for dual-arm aerial manipulator systems.

\end{document}